\documentclass[10pt]{article} 
\usepackage[preprint]{tmlr}

\usepackage{amsmath,amsfonts,bm}

\def\eqref#1{equation~\ref{#1}}

\def\1{\bm{1}}

\DeclareMathAlphabet{\mathsfit}{\encodingdefault}{\sfdefault}{m}{sl}
\SetMathAlphabet{\mathsfit}{bold}{\encodingdefault}{\sfdefault}{bx}{n}

\usepackage{hyperref}
\usepackage{url}
\usepackage{graphicx}
\usepackage{booktabs}
\usepackage[dvipsnames]{xcolor}
\usepackage{array}

\title{Forget Less, Retain More: A Lightweight Regularizer for \\Rehearsal-Based Continual Learning}

\author{
\name Lama Alssum\textsuperscript{*}\quad
      Hasan Abed Al Kader Hammoud\quad
      Motasem Alfarra\quad
      Juan C Leon Alcazar\quad
      Bernard Ghanem\\[0.25em]
      \addr King Abdullah University of Science and Technology (KAUST)\\
      \addr \textsuperscript{*}Corresponding author:
      lama.alssum.1@kaust.edu.sa
}

\def\month{MM}  
\def\year{YYYY} 
\def\openreview{\url{https://openreview.net/forum?id=XXXX}} 

\begin{document}

\maketitle

\begin{abstract}
Deep neural networks suffer from catastrophic forgetting, where performance on previous tasks degrades after training on a new task. This issue arises due to the model’s tendency to overwrite previously acquired knowledge with new information. We present a novel approach to address this challenge, focusing on the intersection of memory-based methods and regularization approaches. We formulate a regularization strategy, termed Information Maximization (IM) regularizer, for memory-based continual learning methods, which is based exclusively on the expected label distribution, thus making it class-agnostic. As a consequence, IM regularizer can be directly integrated into various rehearsal-based continual learning methods, reducing forgetting and favoring faster convergence. Our empirical validation shows that, across datasets and regardless of the number of tasks, our proposed regularization strategy consistently improves baseline performance at the expense of a minimal computational overhead. The lightweight nature of IM ensures that it remains a practical and scalable solution, making it applicable to real-world continual learning scenarios where efficiency is paramount. Finally, we demonstrate the data-agnostic nature of our regularizer by applying it to video data, which presents additional challenges due to its temporal structure and higher memory requirements. Despite the significant domain gap, our experiments show that IM regularizer also improves the performance of video continual learning methods. 
\end{abstract}

\section{Introduction}
\label{sec:intro}

Continual learning (CL) aims to develop models that can learn from evolving data distributions with minimal forgetting \cite{er}. Due to the high computational and financial costs of training deep neural network models and growing concerns over privacy regulations, the applicability of CL in various real-world scenarios has become increasingly critical. For instance, video-sharing platforms such as YouTube and TikTok receive millions of newly uploaded videos daily, each introducing new trends, visual concepts, and styles. In these dynamic environments, traditional training algorithms for deep learning models struggle to maintain their performance due to the necessity of frequent retraining, which is resource-intensive and impractical at scale. CL can significantly enhance the effectiveness of models designed for such dynamic data streams by continually adapting previously trained models, rather than retraining them from scratch as new data is made available.

In recent years, memory-based methods \cite{er,der}, also known as rehearsal methods, have emerged as the front-runners in CL, demonstrating better performance at mitigating forgetting compared to their regularization-based counterparts. This superior performance of rehearsal methods is attributed to the use of a memory buffer, a dedicated storage that retains a subset of training data from previously learned tasks. By having access to a subset of past samples, the model can estimate class prototypes effectively thus alleviate forgetting despite distribution shifts \cite{er, icarl}. This ability to retain past information enables rehearsal-based approaches to maintain stability in long-term learning while adapting to new tasks.

The performance improvement of rehearsal methods over regularization methods comes at the cost of increased memory requirements and greater computational overhead. Experience replay methods retrain for several epochs on both the current data and memory buffer samples, effectively approximating a joint distribution whenever new data becomes available. This continuous reprocessing of stored data samples not only increases computational demands but also increases training time, making it less scalable for large datasets.This computational penalty is further emphasized outside the image domain; for example, in video data, a single minute-long video recorded at 30 frames per second occupies as much memory as 1800 individual images. Despite consuming a significant storage, such video typicality represents a single instance of a class within the memory buffer, thus limiting the diversity of stored information and further challenging the efficiency of experience replay in temporal data. 

In this paper, we introduce a class-agnostic regularization strategy for continual learning (CL), termed Information Maximization (IM), which operates over the distribution of network predictions \cite{shot}. IM is lightweight, efficient, and orthogonal to key design choices in CL algorithms, as it works across different rehearsal-based methods, memory budgets, and number of tasks. IM jointly encourages prediction diversity and confidence, resulting in robust feature representations under distribution shifts, thereby improving generalization to past tasks while minimizing memory and computational overhead. Extensive empirical evaluation shows that IM emerges as an effective regularization technique, outperforming current regularization strategies tailored for the CL setting in both accuracy and retention of past knowledge. Moreover, IM is not specific to the image CL domain; we further validate its effectiveness by applying it to a video continual learning setup, where it demonstrates similarly improved results in handling the challenges posed by temporal dependencies and increased data complexity.

\noindent\textbf{Contributions.} Our work makes three main contributions:
\textbf{(i)} We provide a systematic evaluation of four major regularization techniques: Elastic Weight Consolidation (EWC), Synaptic Intelligence (SI), Entropy Minimization (EM), and Information Maximization (IM), applied to image continual learning. This evaluation highlights the advantages of the proposed IM regularizer, demonstrating its superiority in terms of both performance and overall reduction in catastrophic forgetting.
\textbf{(ii)} We demonstrate that IM is orthogonal to critical design choices in CL (rehearsal-based method, memory size, and number of tasks), enabling straightforward integration while preserving computational efficiency.
\textbf{(iii)} We extend our analysis beyond image-based settings by demonstrating the applicability of IM within the context of video continual learning. Given the additional complexity of temporal dependencies and larger data volumes in videos, our results show that IM maintains its effectiveness, achieving substantial gains over traditional memory-based baselines while preserving computational efficiency.
\section{Related Work}\label{sec:relatedwork}

\paragraph{\textbf{Image Continual Learning.}}

In the field of image-based continual learning, numerous innovative approaches have been proposed to address catastrophic forgetting. Memory-based methods, such as iCaRL \cite{icarl}, utilize incremental classifiers and representation learning to balance new and old knowledge, while GEM \cite{gem} and its more efficient variant A-GEM \cite{a-gem} optimize gradient-based episodic memory to mitigate forgetting. Other approaches, including DER \cite{der}, enhance rehearsal by incorporating logit distillation, while CoPE \cite{de2021continual} leverages class prototypes to structure the latent space, and ER-ACE \cite{caccia2021new} modifies cross-entropy loss to address task imbalance. Recent work includes Refresh Learning \cite{wang2024unified}, which unifies rehearsal with selective unlearning to refresh model knowledge, and STAR \cite{eskandar2025star}, a plug-and-play regularizer that leverages stability-inducing weight perturbations during rehearsal to mitigate forgetting. Regularization-based methods aim to preserve past knowledge by constraining weight updates, typically by identifying the importance of parameters, like Elastic Weight Consolidation (EWC) \cite{kirkpatrick2017overcoming} and Synaptic Intelligence (SI) \cite{si}. Architectural innovations also play a crucial role in continual learning, with L2P \cite{l2p} demonstrating the effectiveness of learnable prompts in guiding pre-trained models without relying on a rehearsal buffer, and DualPrompt \cite{wang2022dualprompt} introducing a two-level prompting mechanism for transformer-based architectures. These diverse approaches underscore the rapid advancements in continual learning, paving the way for more scalable and adaptable models in real-world applications.

\paragraph{\textbf{Video Continual Learning.}}
To mitigate catastrophic forgetting in video data, various strategies have been developed, which can be broadly categorized into regularization and memory-based techniques. While regularization methods apply constraints to preserve previous knowledge, memory-based approaches leverage data or representations from past tasks. When analyzing video continual learning, the importance of memory becomes even more pronounced due to the temporal complexity and higher dimensionality of video data. SMILE \cite{smile} underscores this by proposing an efficient replay mechanism that stores a single frame per video, emphasizing video diversity over temporal information. This approach addresses memory constraints effectively, showcasing the critical role of memory in video continual learning. vCLIMB \cite{vclimb} and PIVOT \cite{pivot} introduce novel benchmarks and methods focusing on class incremental learning and the use of prompting mechanisms, respectively, pushing the boundaries of current methodologies. Utilizing Winning Subnetworks for efficient learning \cite{kang2023continual}, and creating multi-modal datasets for egocentric activity recognition \cite{xu2023towards} illustrate the expanding scope of continual learning in video domains. Additionally, Continual Predictive Learning \cite{chen2022continual} and approaches to Video Object Segmentation as a continual learning task \cite{nazemi2023clvos23} represent significant advancements in handling non-stationary environments and long video sequences. Finally, efforts to learn new class representations while preserving old ones through time-channel importance maps \cite{park2021class} further demonstrate the innovative and diverse strategies being developed for video continual learning.

\paragraph{\textbf{Test-Time Adaptation.}}
Test-Time Adaptation~(TTA) aims to alleviate performance drop of pretrained models at test time when exposed to domain shifts~\cite{ttt, alfarra2023revisiting}.
Earlier works augmented the training objective with a self-supervised loss function that is later leveraged at test time to combat domain shifts~\cite{ttt, ttt++}.
More recent TTA methods optimize an unsupervised loss function at test-time on the received unlabeled data to improve performance under domain shifts~\cite{eata}.
This includes simple adjustments to the statistics of normalization layers~\cite{adabn}, entropy minimization~\cite{wang2020tent}, information maximization~\cite{shot}, among others~\cite{lame, cotta}.
However, most TTA methods are proposed to combat covariate domain shifts at test time.
In this work, we get inspiration from the source hypothesis adaptation method~\cite{shot} to propose an effective regularizer for continual learning.
We also analyze the effectiveness of other adaptation methods such as entropy minimization in mitigating catastrophic forgetting in continual learning.

This work aims to enhance continual learning performance by introducing a cost-effective regularizer that improves results even in memory-constrained scenarios. Such scenarios are particularly important when dealing with memory-intensive data, such as videos, or when sample storage is restricted due to privacy concerns. We investigate a class-independent regularizer designed to facilitate the learning of generalizable features.

\section{Methodology}\label{sec:method}

In this section, we formalize the problem of class-incremental learning in visual recognition tasks. We define the underlying framework and introduce the necessary notation to describe the incremental learning process. Additionally, we present the formulation of the proposed regularizer, Information Maximization (IM), along with the selected baseline regularizers: Elastic Weight Consolidation (EWC), Synaptic Intelligence (SI), and Entropy Minimization (EM).

We focus on the offline continual learning problem for visual recognition tasks, where a classifier $f_\theta:\mathcal X \rightarrow \mathcal P(\mathcal Y)$ (a DNN parameterized by $\theta$) maps an input $x\in\mathcal X$ into the probability simplex\footnote{\emph{e.g.} the network's output after Softmax.} $\mathcal P(\mathcal Y)$ , with $\mathcal Y=\{1, 2, \dots, K\}$. In continual learning, $f_\theta$ is presented with a sequence of $T$ tasks $\{(X_1, Y_1), (X_2, Y_2), \dots , (X_T, Y_T) \}$ where $X_i\subset \mathcal X$ and $Y_i\subset \mathcal Y$ $\forall i$~\cite{vclimb}. Furthermore, we consider the class-incremental problem setup, where the labels presented in each individual task are mutually exclusive ($Y_i \cap Y_j = \phi \,\,\forall i \neq j$). The main objective of the learner is to maximize its performance (\emph{e.g.} classification accuracy) on all observed tasks.
This objective is often hindered by the catastrophic forgetting problem: while learning task $i$, $f_\theta$ tends to forget previously learned tasks $<i$, significantly dropping its performance for any $x \in X_{<i}$.

For our baseline, we consider rehearsal-based continual learning methods where the learner is allowed to store up  to $N$ training examples from previous tasks into a replay memory buffer $M$~\cite{chaudhry2019tiny}. Let $M_t$ denote the replay buffer at task $t$ containing examples from the tasks $i<t$. Rehearsal-based methods update the parameter set $\theta$ at task $t$ in the following form:

\begin{equation}
    \label{eq:rehearsal}
    \theta_t^* =  \underset{\theta}{\arg\min} ~~\mathbb E_{(x, y)\sim(X_t, Y_t)}\mathcal L(f_{\theta}(x), y) + \mathbb E_{(u, v)\sim M_t}\mathcal L(f_{\theta}(u), v).
\end{equation}

That is, for each batch sampled from the newly available data on the $t^{th}$ task, the learner samples another batch from memory $\mathcal{M}_t$ and updates the model on the combined loss.

\subsection{Regularizing Replay Methods with Information Maximization}

Inspired by the work of \cite{shot} in the domain of test-time adaptation, we propose that in a continual learning setup, $f_\theta$ should output confident predictions that distinctly separate all previously seen classes. To achieve this, we propose a regularizer that encourages the model to make confident predictions across all encountered classes without biasing its predictions towards recent task data. This means that for any given input (including those in $\mathcal{M}_t$), the model should assign a high probability to a single class. We can achieve this by maximizing information in the logits, as a result our approach helps reinforce discriminative representations for all learned classes, improving robustness against distribution shifts. Our proposed regularizer ($\mathcal R_{\text{IM}}$) takes the following form:

\begin{equation}
    \label{eq:mutual_information}
    \mathcal R_{\text{IM}}(\theta, X_t) = \mathcal L_{\text{ent}}(\theta, X_t) + \mathcal L_{\text{div}}(\theta, X_t)
\end{equation}
\begin{equation*}
    \text{with }\,\,\mathcal L_{\text{ent}}(\theta, X_t) = -\mathbb{E}_{x\sim X_t}\sum_{k=1}^K f^k_{\theta}(x)\log f^k_{\theta}(x) \qquad \mathcal L_{\text{div}} = \sum_{k=1}^K \hat f^k_{\theta}(x)\log \hat f^k_{\theta}(x),
\end{equation*}

\noindent where $\hat f_{\theta}(x) = \mathbb E_{x\sim X_t} [f_\theta(x)]$ and $f^k_\theta(x)$ is the $k^{th}$ element in the vector $f_\theta(x)$. Note that optimizing $\mathcal L_{\text{ent}}$ increases the model's confidence on the prediction, while $\mathcal L_{\text{div}}$ promotes diverse label predictions on $f_\theta$. Our regularized rehearsal-based method follows the formulation:

\begin{equation}
    \label{eq:regularizing_memory_based_methods}
    \min_\theta ~\mathbb E_{(x, y)\sim(X_t, Y_t)}\mathcal L(f_{\theta}(x), y) + \mathbb E_{(u, v)\sim M_t}\mathcal L(f_{\theta}(u), v) + \mathcal R_{\text{IM}}(\theta, X_t).
\end{equation}

Our proposed regularizer has the following advantages: \textbf{(i)} It is orthogonal to the most critical design choices of continual learning algorithms, as it can operate regardless of the choice of $f_\theta$, the replay-based method, the size of the memory buffer, and the number of tasks. \textbf{(ii)} Efficient computation of $\mathcal R_{IM}$: where both $\mathcal L_{\text{ent}}$ and $\mathcal L_{\text{div}}$ depend exclusively on the output predictions of the model and can be computed in $\mathcal O(n)$. This aspect is essential when dealing with memory-intensive setups. For example, on video data, our regularizer estimates $\mathcal L_{\text{ent}}$ and $\mathcal  L_{\text{div}}$ over clip predictions instead of per-frame estimates. \textbf{(iii)} Our formulation is agnostic to the type of data used in the continual learning problem. Without any modifications, our formulation can be applied to both image-based or video-based continual learning problems.

\subsection{Baseline Regularizers}
We compare our proposal against different regularizers to assess its effectiveness in mitigating forgetting and improving continual learning performance. We follow the formulation in Equation (\ref{eq:regularizing_memory_based_methods}), and study alternatives to $\mathcal R_{\text{IM}}(\theta, X_t)$. In particular, we analyze different regularizers from the continual learning literature, namely Elastic Weight Consolidation~\cite{kirkpatrick2017overcoming} and Synaptic Intelligence~\cite{si}. Furthermore, we explore Entropy Minimization~\cite{wang2020tent} from the test-time adaptation literature.

\paragraph{\textbf{Elastic Weight Consolidation~(EWC).}}
\cite{kirkpatrick2017overcoming} proposed to regularize the parameter update during continual learning to prevent catastrophic forgetting by constraining changes to important weights. The key idea behind EWC is to estimate the importance of each parameter for previously learned tasks and penalize deviations from their learned values. We analyze the effectiveness of combining EWC with rehearsal-based methods by replacing $\mathcal R_{\text{IM}}$ in Equation (\ref{eq:regularizing_memory_based_methods}) with $\mathcal R_{\text{EWC}}$, defined as:

\begin{equation*}
    \mathcal R_{\text{EWC}}(\theta) = \sum_i \frac{\lambda}{2}F_i (\theta^i - \theta_{t-1}^i)^2,
\end{equation*}
where $F$ is the Fisher information matrix, which quantifies the importance of each parameter based on how sensitive the loss function is to changes in that parameter, and $\lambda$ is a hyper-parameter balancing the relative importance of the old tasks with respect to the current task.

\paragraph{\textbf{Synaptic Intelligence (SI).}} It is a biologically inspired regularizer from the continual learning literature. It follows a similar principle to EWC but determines weight importance using a different approach. Instead of using the Fisher Information Matrix, SI tracks the contribution of each parameter during training by accumulating an importance measure based on changes in loss. This adaptive tracking mechanism allows the model to selectively constrain updates to crucial parameters while remaining flexible for learning new tasks. We replace $\mathcal R_{\text{IM}}$ in Equation (\ref{eq:regularizing_memory_based_methods}) with $\mathcal R_{\text{SI}}$ which takes the following form:
\begin{equation*}
  \mathcal R_{\text{SI}}(\theta) = \sum^{T}_{t} \frac{\omega^{k}_{t}}{ (\Delta \theta^{t}_{k})^2 + \xi },
\end{equation*}

\noindent where $\Delta \theta^{t}_{k} = \theta^{t}_{k} - \theta^{t-1}_{k} $ and the damping parameter  $\xi$  avoids division by zero.

\paragraph{\textbf{Entropy Minimization (EM).}} Following the self-supervised spirit of our proposed regularization approach, we include one self-supervised regularizer that encourages the model to produce more confident predictions. In particular, we follow  \cite{wang2020tent} and apply entropy minimization to regularize the output distribution, reducing the model's uncertainty when making predictions.
Entropy minimization replaces $\mathcal R_{\text{IM}}$ in Equation (\ref{eq:regularizing_memory_based_methods}) with $\mathcal R_{\text{EM}}$ where:
 
\begin{equation}
    \label{eq:entropy_minimization}
     \mathcal R_{\text{EM}}(\theta, X_t) = -\mathbb{E}_{x\sim X_t}\sum_{k=1}^K f^k_{\theta}(x)\log f^k_{\theta}(x).
\end{equation}
Entropy minimization encourages the model to assign higher confidence to its predictions, effectively suppressing uncertain outputs. This can be beneficial in a continual learning setup, where distribution shifts can lead to increased uncertainty.
\section{Experiments}\label{sec:experiments}

In this section, we proceed with the empirical assessment of our proposed approach to validate its effectiveness. For completeness, we first evaluate several rehearsal-based continual learning (CL) methods (ER, DER, and DER++) when paired with the regularizers IM, EWC, SI, and EM. We then extend the analysis by applying IM to more advanced rehearsal approaches (Refresh Learning and STAR), showing that its benefits generalize consistently across different methods and datasets. 

\paragraph{\textbf{Datasets.}} Following the image CL literature, we focus on two main datasets: Split-CIFAR100 \cite{si} and Split-Tiny ImageNet \cite{le2015tiny}. Split-CIFAR100 contains a total of 100 classes and 6000 images per class. It is divided into 10 tasks, each containing 10 classes. Split-Tiny ImageNet consists of 200 classes with 500 images per class, and is divided into 10 tasks of 20 classes each. 

\paragraph{\textbf{Evaluation Metrics.}} To evaluate the performance of CL methods, we consider two metrics: Average Accuracy, which is defined as the average performance across all tasks, and Forgetting Rate that measures the impact of the learned task on the performance of the previous tasks \cite{a-gem}. These metrics provide complementary perspectives on the effectiveness of each approach in balancing stability and adaptability in a continual learning setting.

\begin{itemize}

\item \textbf{Average Accuracy (ACC)} quantifies the model's overall performance across all tasks it has encountered. It is defined as:
\begin{equation}
ACC = \frac{1}{T} \sum_{i=1}^{T} a_i, 
\end{equation}
where \(T\) is the total number of tasks and \(a_i\) represents the accuracy of the model on the \(i\)-th task after it has been trained on all \(T\) tasks. ACC provides a comprehensive measure of how well the model learns and retains knowledge across a full sequence of tasks.
~\\
\item \textbf{Forgetting Rate (FR)} measures the decrease in performance on past tasks after a model has been trained on new ones. It directly measures catastrophic forgetting. FR is defined as:
\begin{equation}
FR = \frac{1}{T-1} \sum_{i=1}^{T-1} \max_{j < T}(a_{ij} - a_{iT}),
\end{equation}
where \(a_{ij}\) is the accuracy on task \(i\) immediately after training on task \(j\), and \(a_{iT}\) is the accuracy on task \(i\) after the final task \(T\) has been learned. A lower FR indicates better retention of previously learned knowledge, while a higher FR points to significant forgetting.
\end{itemize}

\paragraph{\textbf{Implementation Details.}} We train a ResNet18 \cite{he2016deep} model from scratch and summarize the training parameters in the \textbf{Appendix}. Following our limited memory setting, we define a budget of 5, 10 and 20 samples per class as the maximum allowed in $\mathcal{M}_{t}$ at any moment. We integrate the different regularizers into the Mammoth CL framework \cite{boschini2022class, buzzega2020dark}. To balance the loss terms, we multiply the regularization term by $\lambda$=0.5 and the cross-entropy loss with 1-$\lambda$.

\begin{figure}[t]
    \centering
   
    \includegraphics[width=\textwidth]{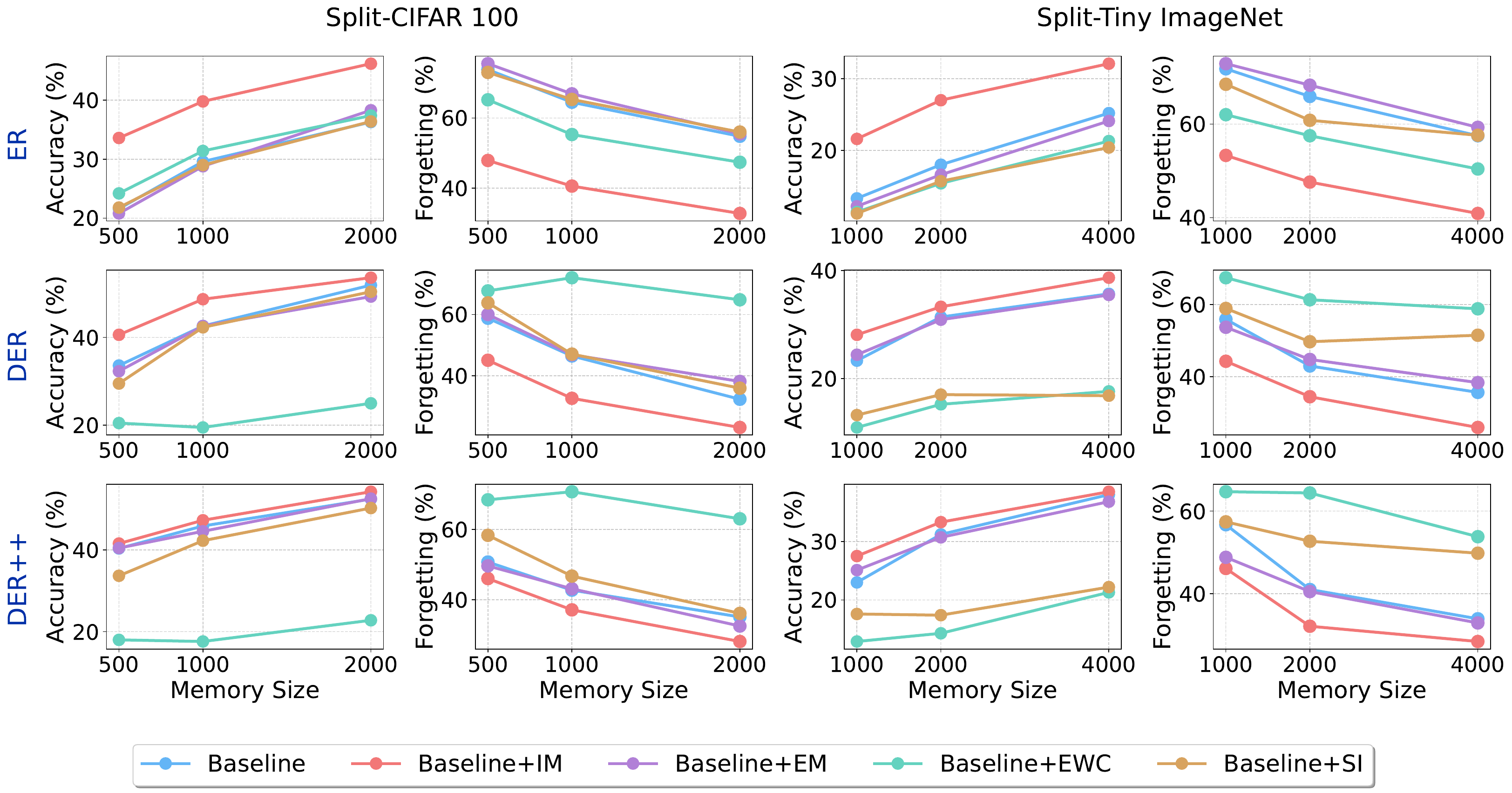}
   \newline
    \caption{ \textbf{Results of Integrating Different Regularizers on Split-CIFAR100 and Split-Tiny
ImageNet. } This figure plots the average accuracy and   forgetting rate of three baseline methods (ER, DER, and DER++) across various sizes of memory buffer, in    combination with the analyzed regularizers (IM, EM, EW, and SI), and across two datasets (Split-CIFAR100 and Split-Tiny ImageNet). The results demonstrate that the proposed information maximization regularizer (IM) consistently outperforms other methods, achieving higher accuracy and lower forgetting rates on both datasets regardless of the memory setting.
     }
    \label{fig:main-results}
\end{figure}

\begin{figure}[t]
    \centering
   
    \includegraphics[width=\textwidth]{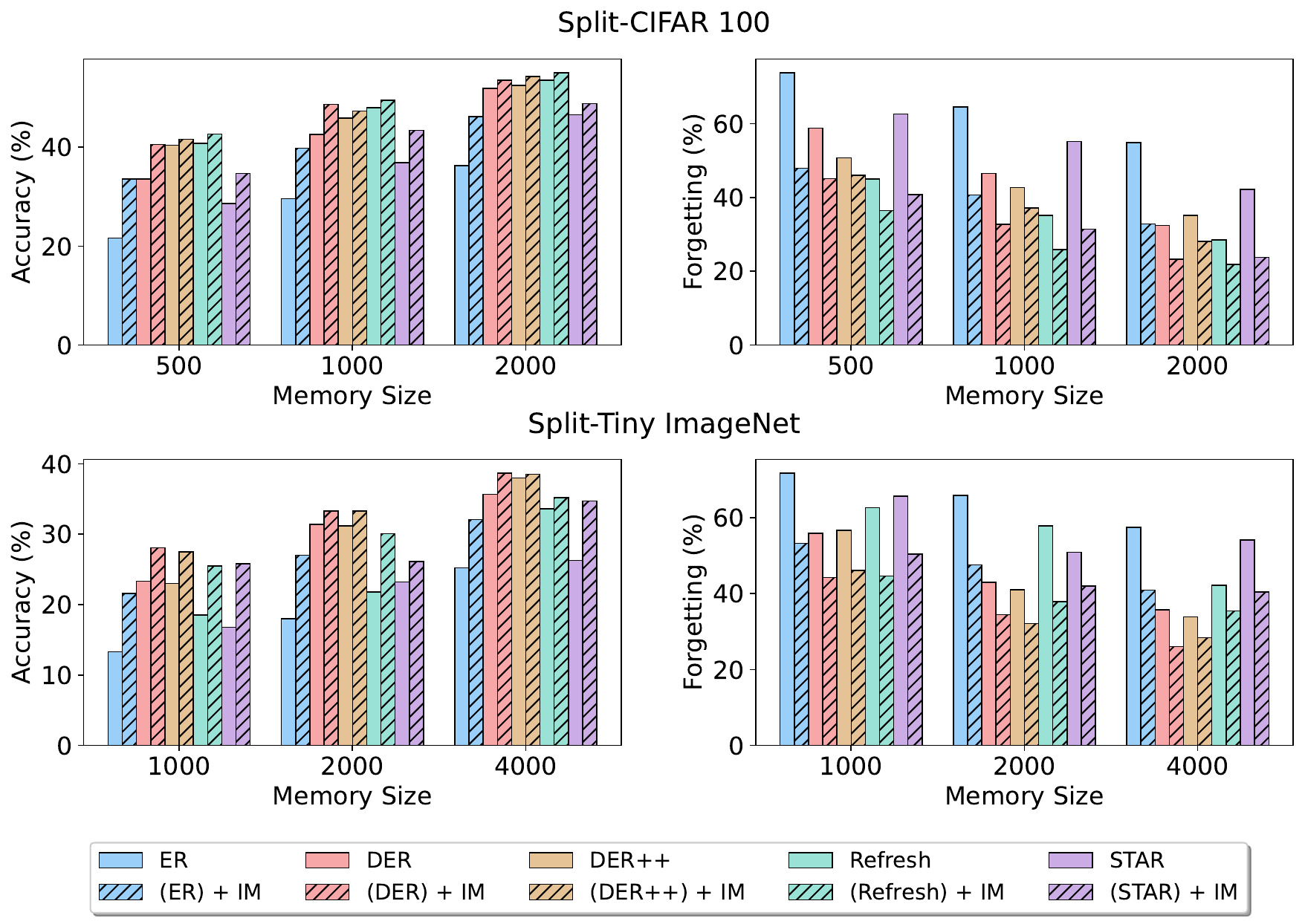}
    \newline
    \caption{ \textbf{Results on Split-CIFAR100 and Split-Tiny ImageNet.} This figure presents the average accuracy and forgetting rate of five rehearsal-based methods (ER, DER, DER++, Refresh Learning, and STAR) across different memory buffer sizes, both in their baseline form and when combined with Information Maximization (IM). The results show that integrating IM consistently enhances performance, leading to higher accuracy and reduced forgetting across all methods, datasets, and memory settings.
     }
    \label{fig:main-results-all}
\end{figure}
\vspace{\baselineskip}

\paragraph{\textbf{Baselines.}} We consider four regularizers: EWC \cite{kirkpatrick2017overcoming}, SI \cite{si}, EM \cite{wang2020tent}, and IM \cite{shot}, applied on top of three memory-based continual learning methods: ER \cite{er}, DER \cite{der}, and DER++ \cite{der}. For the IM regularizer, we further extend the evaluation to include Refresh Learning \cite{wang2024unified}, implemented on top of DER++, 
and STAR \cite{eskandar2025star}, implemented on top of ER.  

\subsection{Regularized Rehearsal Methods Results} 
\label{subsec:MainResults}
Figure (\ref{fig:main-results}) summarizes the performance of rehearsal-based methods ER, DER, and DER++ on Split-CIFAR100 (left columns) and Split-Tiny ImageNet (right columns) for the selected memory sizes. We include the baseline performance (light blue) and outline the impact of incorporating regularization techniques on top of these rehearsal methods. 

Our results demonstrate that introducing IM on top of rehearsal-based methods consistently leads to improvements across all memory sizes. For instance, when IM is applied to ER on Split-CIFAR100, we observe an enhancement of \textcolor{ForestGreen}{10-13\%} in performance across all memory sizes. The improvement is slightly lower for DER and DER++, ranging around \textcolor{ForestGreen}{2-7\%} for DER and \textcolor{ForestGreen}{1-2\%} for DER++. In contrast, other regularizers like EWC, SI, and EM generally do not improve the baseline methods, and can even degrade performance in some cases, as is the case for EWC on DER and DER++, where the model's accuracy drops by nearly half. 

We can also observe in Figure (\ref{fig:main-results}) that forgetting is significantly reduced when the rehearsal methods are paired with IM. For ER (paired with IM) applied on Split-CIFAR100, the reduction in forgetting is around (22-25\%) across all memory sizes. For DER and DER++ on Split-CIFAR100, our results show a smaller reduction compared to ER with about (9-13\%) and (4-6\%), respectively. On the other hand, EM and SI do not generally reduce forgetting compared to the original baselines. 
This is since EM aims at increasing the model's confidence in predicting samples from the current task. While this approach might accelerate the learning process over tasks, it does not promote the retention of previously learned information. 

For EWC, we observe reduction in forgetting when paired with ER, but not with the remaining baselines. More detailed and comprehensive results can be found in the \textbf{Appendix}. 

To further validate the effectiveness of IM, we expand our analysis to include two additional rehearsal-based baselines: Refresh Learning and STAR. Figure (\ref{fig:main-results-all}) presents the performance gains when IM is integrated into all five rehearsal methods (ER, DER, DER++, Refresh Learning, STAR) across both Split-CIFAR100 and Split-Tiny ImageNet. 

On Split-CIFAR100, IM consistently improves the performance of Refresh Learning and STAR, but with varying impact. When paired with Refresh Learning, the improvements are more modest, about \textcolor{ForestGreen}{1–2\%} in accuracy, but forgetting is reduced by (7–9\%) across all memory sizes. This indicates that while Refresh Learning already stabilizes training to some degree, IM provides an additional layer of retention without significantly altering the learning dynamics. For STAR, the effect is more significant, as we observe accuracy gains of \textcolor{ForestGreen}{2–6\%}, along with a substantial reduction in forgetting (18–24\%). 

On Split-Tiny ImageNet, IM continues to yield consistent improvements. For Refresh Learning, the accuracy gains are larger than on Split-CIFAR100, ranging from \textcolor{ForestGreen}{2–8\%}, with forgetting reduced by (7–20\%). STAR also benefits considerably, with accuracy improvements of \textcolor{ForestGreen}{3–9\%} and forgetting reductions between (9–15\%). 


\paragraph{\textbf{Conclusion}.} These results reveal that Information Maximization (IM) serves as an effective regularization technique, consistently improving the performance of image continual learning baselines across various memory budgets. By encouraging confident yet balanced predictions, IM enhances both accuracy and knowledge retention, thereby mitigating catastrophic forgetting.

\subsection{Ablation Analysis}
To explore the limitations of Information Maximization (IM) as a regularizer for continual learning methods, we conduct three ablation experiments aimed at understanding its performance under various conditions. First, we assess the impact of IM when the computational budget is reduced to determine whether it improves the convergence of the baseline methods. Second, we evaluate if the improvement obtained by using IM diminishes with additional tasks. Specifically, we investigate the performance of our approach when the number of tasks exceeds 10, to understand how IM scales with an increasing number of tasks. Third, we examine the effect of applying IM exclusively to buffer samples, as opposed to both buffer and current task samples, to assess whether applying IM beyond current task samples is beneficial.

\paragraph{\textbf{Computational Budget}.}

In the experiments presented in Section (\ref{subsec:MainResults}), the computational budget was set following the default settings of Mammoth CL framework, \emph{i.e.} 50 epochs per task for Split-CIFAR100 and 100 epochs per task for Split-Tiny ImageNet. These settings ensure sufficient training iterations for each task, allowing models to stabilize and learn meaningful representations. However, in practical scenarios, computational efficiency is a critical factor, as training deep learning models over multiple tasks can be prohibitively expensive. Currently, there is a growing interest in exploring low-computational regimes \cite{ghunaim2023real, prabhu2023computationally} for CL, as computing resources are far more expensive than storage \cite{prabhu2023computationally}. This shift in focus reflects the necessity of developing continual learning approaches that remain effective even with reduced training time. Therefore, to investigate how our proposed regularizer, IM, performs under computational constraints, we run ablation experiments with about half of the computational budget on Split-Tiny ImageNet (\textit{i.e.} half of the training epochs per task) and less than a quarter of the computational budget on Split-CIFAR100 compared to our original experimental setup.

The results in Table (\ref{tab:compute_budget}) show that, even with a lower computational budget of 10 and 50 epochs per task on Split-CIFAR100 and Split-Tiny ImageNet, respectively, the proposed ER+IM and DER+IM methods outperform their counterparts without IM regularizer. For instance, on the Split-CIFAR100 dataset with a buffer size of 1000, ER+IM achieves an accuracy of 35.7\%, significantly higher than ER at 26.8\%. Similarly, DER+IM attains 37.3\% accuracy, surpassing DER's 21.1\% by a large margin. These trends hold for different buffer sizes and datasets, highlighting the effectiveness of the proposed regularization in low-compute regimes.  

\begin{table}[t]
    \centering
        \caption{
\textbf{Ablation Study on Compute Budget.}
This table presents the performance of ER and DER, with and without the Information Maximization (IM) regularizer, on Split-CIFAR100 and Split-Tiny ImageNet datasets. For Split-CIFAR100, the baselines are run with 10 epochs instead of the original 50, while for Split-Tiny ImageNet, the experiments are run with 50 epochs instead of the original 100. The results indicate that incorporating IM leads to improved convergence and consistently higher average accuracy.}\label{tab:compute_budget}
           \vspace{\baselineskip}
            \begin{tabular}{l@{\hskip 0.5cm}ccc@{\hskip 0.5cm}ccc}
            
                \toprule
                & \multicolumn{3}{c}{\textbf{Split-Cifar100}} & \multicolumn{3}{c}{\textbf{Split-Tiny ImageNet}} \\
                \cmidrule(r){2-4} \cmidrule(l){5-7}
                \textbf{Buffer Size} & \textbf{500} & \textbf{1000} & \textbf{2000} & \textbf{1000} & \textbf{2000} & \textbf{4000} \\
                \midrule
                ER & 20.8 & 26.8 & 35.6 & 13.2 & 18.7 & 25.6 \\
                ER + IM &\textbf{28.8} & \textbf{35.7} & \textbf{40.1} & \textbf{21.5} & \textbf{26.6} & \textbf{31.8} \\ \midrule
                DER & 24.1 & 21.1 & 19.9 & 21.6 & 26.3 & 24.2 \\
                DER + IM & \textbf{33.5} & \textbf{37.3} & \textbf{34.4} & \textbf{27.9} & \textbf{32.1} & \textbf{33.3} \\
                \bottomrule
            \end{tabular}
        
    \end{table}

\paragraph{\textbf{Number of Tasks}.}
In the experiments presented in Section (\ref{subsec:MainResults}), we used the conventional 10-tasks split for Split-CIFAR100 and Split-Tiny ImageNet, which is commonly used in continual learning studies. However, as shown in \cite{vclimb,prabhu2023computationally}, performance may vary when more tasks are introduced. More tasks can make the problem harder because the model has to remember more information and avoid forgetting earlier tasks while learning new information. Consequently, we reran the experiments in Section (\ref{subsec:MainResults}), doubling the number of tasks from 10 to 20. This allows us to evaluate whether Information Maximization (IM) regularizer remains effective when the continual learning problem becomes more challenging due to having more tasks to learn. The results presented in Table (\ref{ref:num_tasks_ablation}) show that incorporating IM into ER and DER can significantly improve their performance on longer sequences of tasks. For example, ER+IM shows an improvement of (4-7\%) and (6-8\%) on Split-CIFAR100 and Split-Tiny ImageNet, respectively. On the other hand, DER+IM shows a (4-7\%) and (4-5\%) improvement on Split-CIFAR100 and Split-Tiny ImageNet, respectively.

\begin{table}[t]
    \centering
    \caption{\textbf{Ablation Study on Number of Tasks.} This table presents the performance of ER and DER , with and without Information Maximization (IM) regularizer, on a sequence of 20 tasks for both Split-CIFAR100 and Split-Tiny ImageNet datasets. Our evaluation shows that, despite the increased amount of tasks our proposal still outperform the ER and DER baselines in every single scenario.}
    \label{ref:num_tasks_ablation}
        \vspace{\baselineskip}
            \begin{tabular}{l@{\hskip 0.5cm}ccc@{\hskip 0.5cm}ccc}
                \toprule
                & \multicolumn{3}{c}{\textbf{Split-CIFAR100}} & \multicolumn{3}{c}{\textbf{Split-Tiny ImageNet}} \\
                \cmidrule(r){2-4} \cmidrule(l){5-7}
                \textbf{Buffer Size} & \textbf{500} & \textbf{1000} & \textbf{2000} & \textbf{1000} & \textbf{2000} & \textbf{4000} \\
                \midrule
                ER &16.6&25.8&34.7&9.1&14.5&22.1\\
                ER + IM &\textbf{23.4}&\textbf{31.3}&\textbf{38.9}&\textbf{18.3}&\textbf{23.4}&\textbf{29.1}\\ \midrule
                DER &25.1&35.8&38.9&18.0&22.5&27.8\\
                DER + IM &\textbf{32.8}&\textbf{39.8}&\textbf{45.0}&\textbf{23.2}&\textbf{27.0}&\textbf{32.1}\\
                \bottomrule
                
            \end{tabular}
        
\end{table}

\paragraph{\textbf{Regularization Targets}.}

For the experimental assessment in Section (\ref{subsec:MainResults}), we apply the regularization loss to current task samples only. This raises the question of how the proposed method would behave if the IM loss were applied exclusively to memory samples or to both memory and current task samples. For this reason, we reran the experiments shown in Section (\ref{subsec:MainResults}) for both variants, and the results are summarized in Table (\ref{tab:target_ablation}). We find that applying the IM loss to the current task (CT) is superior to applying it to the memory/buffer samples only (BF) or to both buffer and current task samples (ALL). Notably, this trend remains consistent across various buffer sizes, datasets, and continual learning methods, highlighting the robustness of this strategy. 

For example, with a buffer size of 500 on the Split-CIFAR100, ER+IM (CT) achieves an accuracy of 33.6\%, which is significantly higher than ER+IM (ALL) and ER+IM (BF), which achieve 25.9\% and 21.7\%, respectively. Similarly, DER+IM (CT) consistently outperforms DER+IM (ALL) and DER+IM (BF) across various buffer sizes and datasets, reinforcing the advantage of applying IM solely to current task samples. For example, it achieves 48.7\% accuracy on Split-CIFAR100 with a buffer size of 1000, while DER +IM (ALL) and DER +IM (BF) achieve 46.0\% and 41.0\%, respectively. These results indicate that applying IM loss to the current task samples is more effective than applying it exclusively to the memory/buffer samples only or to both buffer and current task samples.

\begin{table}[t]
        \centering
\vspace{\baselineskip}
\caption{
\textbf{Ablation Study on Regularization Target Selection.}
In the main results, the Information Maximization (IM) regularizer is applied exclusively to the current task samples (CT). This table presents the results of applying the regularizer to the buffer samples only (BF) and to both current task samples and buffer samples simultaneously (ALL). The findings indicate that applying the regularizer to the current task samples consistently leads to superior performance compared to the other variants.
}\label{tab:target_ablation}
            \vspace{\baselineskip}
            \begin{tabular}{l@{\hskip 0.5cm}ccc@{\hskip 0.5cm}ccc}

                \toprule
                & \multicolumn{3}{c}{\textbf{Split-CIFAR100}} & \multicolumn{3}{c}{\textbf{Split-Tiny ImageNet}} \\
                \cmidrule(r){2-4} \cmidrule(l){5-7}
                \textbf{Buffer Size} & \textbf{500} & \textbf{1000} & \textbf{2000} & \textbf{1000} & \textbf{2000} & \textbf{4000} \\
                \midrule
                ER + IM (ALL) & 25.9 & 34.6 & 42.7 & 15.0 & 20.3 & 27.8 \\
                ER + IM (CT) & \textbf{33.6} & \textbf{39.8} & \textbf{46.2} & \textbf{21.6} & \textbf{27.0} & \textbf{32.1} \\
                ER + IM (BF) & 21.7 & 28.9 & 38.8 & 12.7 & 17.7 & 24.2 \\ \midrule
                DER + IM (ALL) & 33.5 & 46.0 & 53.5 & 27.7 & 32.7 & 35.1 \\
                DER + IM (CT) & \textbf{40.6} & \textbf{48.7} & \textbf{53.6} & \textbf{28.1} & \textbf{33.3} & \textbf{38.7} \\
                DER + IM (BF) & 27.3 & 41.0 & 50.2 & 21.2 & 27.7 & 34.6 \\
                \bottomrule
                
            \end{tabular}
\vspace{\baselineskip}   
\end{table}

\subsection{Generalization to Video Continual Learning}

To further validate the effectiveness of Information Maximization (IM) as a cost-effective regularization technique for enhancing continual learning methods, we extend our analysis to representative experiments in the video domain. Specifically, we experiment with the iCaRL approach as part of the popular vCLIMB framework \cite{icarl,vclimb} to assess IM's performance in this video CL context.

We evaluate the use of IM loss on two widely recognized datasets in the video domain: UCF-101, consisting of 101 classes split into 10 tasks, and ActivityNet, comprising 200 classes also divided into 10 tasks. We set the memory size to 5, 10 and 20 samples per class and adopt the training hyperparameters from \cite{vclimb}.

\begin{figure}[t]
    \centering

    \includegraphics[width=\textwidth]{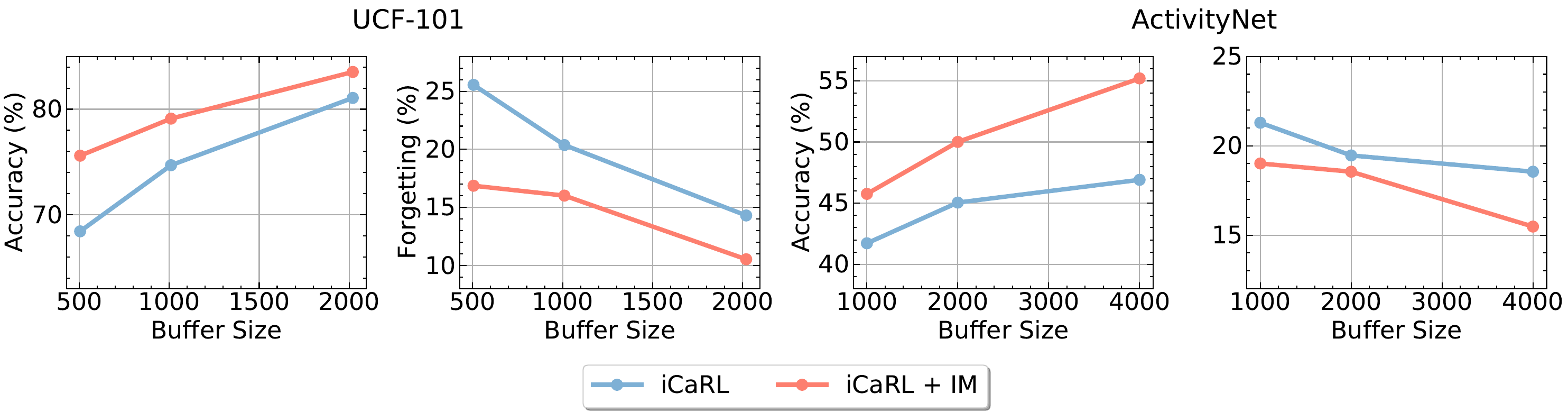}
    \newline
        \caption{
\textbf{Application of Information Maximization to Video Continual Learning.}
This figure illustrates the average accuracy and forgetting rates of the iCARL video continual learning variant, introduced by vCLIMB \cite{vclimb}, with and without our Information Maximization (IM) regularizer. The results demonstrate that incorporating the IM regularizer on top of iCARL leads to consistent improvements in average accuracy and reductions in the forgetting rate.
}

\label{fig:video_cl}
\end{figure}

Upon applying iCARL to UCF-101 with IM, we achieve an improvement of \textcolor{ForestGreen}{2-8\%} across all memory sizes. Similarly, on ActivityNet, the accuracy gain is between \textcolor{ForestGreen}{4-8\%}  with the incorporation of IM. These results further underscore the potential of IM as a valuable regularization technique for enhancing performance in the video domain and continual learning scenarios. Note that video data has an additional temporal dimension compared to images, which requires more memory to store. Being able to improve continual learning performance with small memory buffer sizes, as shown in Figure \ref{fig:video_cl}, is crucial for facilitating the development of memory-efficient approaches for video continual learning.

\vspace{\baselineskip}
\section{Conclusion}\label{sec:conclusion}
In conclusion, this paper explores the combined potential of memory-based methods and regularization techniques in the context of Continual Learning (CL), specifically within a class incremental setup. We introduce a novel, class-agnostic regularization strategy for CL, which focuses on the distribution of the network’s predictions. This strategy, termed Information Maximization (IM) regularization, facilitates the learning of enhanced feature representations across multiple distribution shifts, while simultaneously minimizing memory requirements and computational overhead. Our extensive empirical evaluation underscores the effectiveness of the proposed IM regularizer, demonstrating its superiority over existing regularization strategies designed for CL. Furthermore, the simplicity and versatility of our approach allow it to be applied across different input domains, as evidenced by its successful application in the video continual learning setup. Unlike traditional image-based settings, video CL presents additional challenges due to its temporal structure and higher memory demands. Despite these complexities, our method demonstrates strong performance, reinforcing its applicability to real-world, resource-constrained scenarios. 
\section*{Acknowledgment}
This work is supported by the KAUST Center of
Excellence for Generative AI under award number
5940. The computational resources are provided
by IBEX, which is managed by the Supercomputing Core Laboratory at KAUST.
\bibliography{main}
\bibliographystyle{tmlr}

\newpage

\section{Appendix}

\subsection{Regularized Rehearsal Methods Results}
Tables \ref{tab:acc_results} and \ref{tab:forgetting_results}, contain the numerical results for the average accuracy and forgetting rate metrics, respectively, on three baseline methods (ER, DER, and DER++) in combination with the analyzed regularizers (IM, EM, EW, and SI), as well as for integrating IM into Refresh Learning and STAR. These results were summarized as plots in Figures (\ref{fig:main-results}) and (\ref{fig:main-results-all}) of the main paper. We observe that our proposed regularize (IM) consistently outperforms other methods across different memory settings. The hyper-parameters we used for every baseline when combined with the analyzed regularizers (IM, EM, EW, and SI) are listed in Table \ref{tab:cifar_parameters} and \ref{tab:tiny_parameters}. 

\begin{table*}[h]
\centering
\vspace{\baselineskip}
\caption{
\textbf{Average Accuracy on Split-CIFAR100 and Split-Tiny ImageNet.} This table shows
the average accuracy of three baseline methods (ER, DER, and
DER++) across various sizes of memory buffer, in combination with the analyzed
regularizers (IM, EM, EW, and SI), as well as for integrating IM into Refresh Learning and STAR. 
}\label{tab:acc_results}
\vspace{\baselineskip}
\begin{tabular}{@{}lcccccc@{}}
\toprule
Dataset    & \multicolumn{3}{c}{Split-CIFAR100}& \multicolumn{3}{c}{Split-Tiny ImageNet}       \\ \midrule
Buffer      & 500  & 1000 & 2000 & 1000 & 2000 & 4000 \\ \midrule
ER          & 21.6 & 29.6 & 36.3 & 13.3 & 18.0 & 25.2 \\
ER (IM)    & \textbf{33.6} & \textbf{39.8} & \textbf{46.2} & \textbf{21.6} & \textbf{27.0} & \textbf{32.1} \\
ER (EM)     & 20.8 & 28.8 & 38.3 & 12.2 & 16.6 & 24.1 \\
ER (EWC)    & 24.2 & 31.4 & 37.4 & 11.4 & 15.4 & 21.3 \\
ER (SI)     & 21.8 & 29.0 & 36.4 & 11.2 & 15.7 & 20.4 \\ \midrule
DER         & 33.6 & 42.6 & 51.9 & 23.3 & 31.4 & 35.7 \\
DER (IM)   & \textbf{40.6} & \textbf{48.7} & \textbf{53.6} & \textbf{28.1} & \textbf{33.3} & \textbf{38.7} \\
DER (EM)    & 32.3 & 42.6 & 49.3 & 24.4 & 30.9 & 35.5 \\
DER (EWC)   & 20.5 & 19.5 & 25.0 & 10.9 & 15.2 & 17.6 \\
DER (SI)    & 29.5 & 42.3 & 50.4 & 13.2 & 17.0 & 16.8 \\ \midrule
DER++       & 40.4 & 45.9 & 52.5 & 23.0 & 31.2 & 38.0 \\
DER++ (IM) & \textbf{41.6} & \textbf{47.3} & \textbf{54.3} & \textbf{27.5} & \textbf{33.3} & \textbf{38.5} \\
DER++ (EM)  & 40.5 & 44.6 & 52.6 & 25.1 & 30.7 & 36.8 \\
DER++ (EWC) & 18.0 & 17.6 & 22.8 & 12.9 & 14.3 & 21.3 \\
DER++ (SI)  & 33.7 & 42.3 & 50.3 & 17.6 & 17.4 & 22.2 \\ \midrule
Refresh     & 40.8 & 48.0 & 53.6 & 18.5 & 21.8 & 33.6 \\
Refresh (IM) & \textbf{42.7} & \textbf{49.5} & \textbf{55.1} & \textbf{25.5} & \textbf{30.1} & \textbf{35.2} \\ \midrule
STAR     & 28.6 & 36.9 & 46.6 & 16.8 & 23.2 & 26.3 \\
STAR (IM) & \textbf{34.7} & \textbf{43.4} & \textbf{48.8} & \textbf{25.8} & \textbf{26.1} & \textbf{34.7} \\
\bottomrule
\end{tabular}
\end{table*}
\vspace{\baselineskip}
\begin{table*}[h]
\centering
\vspace{\baselineskip}
\caption{
\textbf{Forgetting Rate on Split-CIFAR100 and Split-Tiny ImageNet.} This table shows
the forgetting rate of three baseline methods (ER, DER, and
DER++) across various sizes of memory buffer, in combination with the analyzed
regularizers (IM, EM, EW, and SI), as well as the results for integrating IM into Refresh Learning and STAR.
}\label{tab:forgetting_results}
\vspace{\baselineskip}
\begin{tabular}{@{}lcccccc@{}}
\toprule
Dataset    & \multicolumn{3}{c}{Split-CIFAR100}& \multicolumn{3}{c}{Split-Tiny ImageNet}       \\ \midrule
Buffer      & 500  & 1000 & 2000 & 1000 & 2000 & 4000 \\ \midrule
ER          & 73.8 & 64.5 & 54.8 & 71.8 & 65.9 & 57.5 \\
ER (IM)    & \textbf{47.9} & \textbf{40.6} & \textbf{32.8} & \textbf{53.3} & \textbf{47.6} & \textbf{40.9} \\
ER (EM)     & 75.5 & 66.9 & 55.5 & 72.9 & 68.3 & 59.3 \\
ER (EWC)    & 65.2 & 55.3 & 47.4 & 62.0 & 57.5 & 50.4 \\
ER (SI)     & 73.0 & 65.3 & 56.0 & 68.5 & 60.8 & 57.6 \\ \midrule
DER         & 58.8 & 46.5 & 32.4 & 55.9 & 43.0 & 35.7 \\
DER (IM)   & \textbf{45.1} & \textbf{32.7} & \textbf{23.2} & \textbf{44.3} & \textbf{34.5} & \textbf{26.0} \\
DER (EM)    & 60.0 & 46.9 & 38.2 & 53.7 & 44.8 & 38.4 \\
DER (EWC)   & 67.7 & 72.0 & 64.8 & 67.4 & 61.3 & 58.8 \\
DER (SI)    & 63.8 & 47.1 & 36.0 & 58.9 & 49.7 & 51.5 \\ \midrule
DER++       & 50.7 & 42.7 & 35.1 & 56.7 & 41.0 & 33.9 \\
DER++ (IM) & \textbf{46.0} & \textbf{37.1} & \textbf{28.1} & \textbf{46.1} & \textbf{32.1} & \textbf{28.4} \\
DER++ (EM)  & 49.6 & 43.1 & 32.5 & 48.8 & 40.5 & 32.9 \\
DER++ (EWC) & 68.4 & 70.7 & 63.0 & 64.7 & 64.4 & 53.8 \\
DER++ (SI)  & 58.3 & 46.7 & 36.1 & 57.4 & 52.7 & 49.8 \\ \midrule
Refresh     & 45.0 & 35.1 & 28.5 & 62.7 & 57.9 & 42.2 \\
Refresh (IM) & \textbf{36.4} & \textbf{25.9} & \textbf{21.8} & \textbf{44.6} & \textbf{37.9} & \textbf{35.4} \\ \midrule
STAR     & 62.6 & 55.1 & 42.2 & 65.7 & 50.9 & 54.1 \\
STAR (IM) & \textbf{40.8} & \textbf{31.4} & \textbf{23.7} & \textbf{50.4} & \textbf{42.0} & \textbf{40.4} \\ \bottomrule
\end{tabular}
\end{table*}
\vspace{\baselineskip}
\begin{table*}[t]
\centering
\renewcommand{\arraystretch}{1.2}
\caption{
\textbf{Experiments Hyper-Parameters (Split-CIFAR100).} 
Columns list the training hyper-parameters. 
Buf. = buffer size, LR = learning rate, Mom. = momentum, WD = weight decay, 
$\alpha$ = distillation coefficient (used in DER/DER++/Refresh), 
$\beta$ = memory balancing coefficient (used in DER++/Refresh), 
$\gamma$, $\lambda$, $s$ (specific to STAR), 
$\Gamma$, $J$ (specific to Refresh Learning). 
A dash (--) indicates that the parameter is not applicable to the method.
}
\label{tab:cifar_parameters}
\begin{tabular}{
    p{0.10\textwidth}
    >{\centering\arraybackslash}p{0.06\textwidth}
    >{\centering\arraybackslash}p{0.06\textwidth}
    >{\centering\arraybackslash}p{0.06\textwidth}
    >{\centering\arraybackslash}p{0.06\textwidth}
    >{\centering\arraybackslash}p{0.05\textwidth}
    >{\centering\arraybackslash}p{0.05\textwidth}
    >{\centering\arraybackslash}p{0.05\textwidth}
    >{\centering\arraybackslash}p{0.05\textwidth}
    >{\centering\arraybackslash}p{0.05\textwidth}
    >{\centering\arraybackslash}p{0.05\textwidth}
    >{\centering\arraybackslash}p{0.05\textwidth}
    >{\centering\arraybackslash}p{0.05\textwidth}
}
\toprule
Baseline & Buf. & LR & Mom. & WD & $\alpha$ & $\beta$ & $\gamma$ & $\lambda$ & $s$ & $\Gamma$ & $J$ \\ 
\midrule
ER       &  500 & 0.1  & 0 & 0 & --  & --  & -- & -- & -- & -- & -- \\
ER       & 1000 & 0.1  & 0 & 0 & --  & --  & -- & -- & -- & -- & -- \\
ER       & 2000 & 0.1  & 0 & 0 & --  & --  & -- & -- & -- & -- & -- \\ \midrule
DER      &  500 & 0.03 & 0 & 0 & 0.3 & --  & -- & -- & -- & -- & -- \\
DER      & 1000 & 0.03 & 0 & 0 & 0.3 & --  & -- & -- & -- & -- & -- \\
DER      & 2000 & 0.03 & 0 & 0 & 0.3 & --  & -- & -- & -- & -- & -- \\ \midrule
DER++    &  500 & 0.03 & 0 & 0 & 0.3 & 0.5 & -- & -- & -- & -- & -- \\
DER++    & 1000 & 0.03 & 0 & 0 & 0.3 & 0.5 & -- & -- & -- & -- & -- \\
DER++    & 2000 & 0.03 & 0 & 0 & 0.3 & 0.5 & -- & -- & -- & -- & -- \\ \midrule
Refresh  &  500 & 0.03 & 0 & 1e-4 & 0.3 & 0.5 & -- & -- & -- & 1e-5 & 1 \\
Refresh  & 1000 & 0.03 & 0 & 1e-4 & 0.3 & 0.5 & -- & -- & -- & 1e-5 & 1 \\
Refresh  & 2000 & 0.03 & 0 & 1e-4 & 0.3 & 0.5 & -- & -- & -- & 1e-5 & 1 \\ \midrule
STAR     &  500 & 0.1  & 0 & 0 & --  & --  & 0.05 & 0.05 & 1 & -- & -- \\
STAR     & 1000 & 0.05 & 0 & 0 & --  & --  & 0.05 & 0.05 & 1 & -- & -- \\
STAR     & 2000 & 0.05 & 0 & 0 & --  & --  & 0.05 & 0.05 & 1 & -- & -- \\
\bottomrule
\end{tabular}
\end{table*}
\vspace{\baselineskip}
\begin{table*}[t]
\centering
\renewcommand{\arraystretch}{1.2}
\caption{
\textbf{Experiments Hyper-Parameters (Split-Tiny ImageNet).} 
Columns list the training hyper-parameters. 
Buf. = buffer size, LR = learning rate, Mom. = momentum, WD = weight decay, 
$\alpha$ = distillation coefficient (used in DER/DER++/Refresh), 
$\beta$ = memory balancing coefficient (used in DER++/Refresh), 
$\gamma$, $\lambda$, $s$ (specific to STAR), 
$\Gamma$, $J$ (specific to Refresh Learning). 
A dash (--) indicates that the parameter is not applicable to the method.
}
\label{tab:tiny_parameters}
\begin{tabular}{
    p{0.10\textwidth}
    >{\centering\arraybackslash}p{0.06\textwidth}
    >{\centering\arraybackslash}p{0.06\textwidth}
    >{\centering\arraybackslash}p{0.06\textwidth}
    >{\centering\arraybackslash}p{0.06\textwidth}
    >{\centering\arraybackslash}p{0.05\textwidth}
    >{\centering\arraybackslash}p{0.05\textwidth}
    >{\centering\arraybackslash}p{0.05\textwidth}
    >{\centering\arraybackslash}p{0.05\textwidth}
    >{\centering\arraybackslash}p{0.05\textwidth}
    >{\centering\arraybackslash}p{0.05\textwidth}
    >{\centering\arraybackslash}p{0.05\textwidth}
    >{\centering\arraybackslash}p{0.05\textwidth}
}
\toprule
Baseline & Buf. & LR & Mom. & WD & $\alpha$ & $\beta$ & $\gamma$ & $\lambda$ & $s$ & $\Gamma$ & $J$ \\ 
\midrule
ER       &  500 & 0.03  & 0 & 0 & --  & --  & -- & -- & -- & -- & -- \\
ER       & 1000 & 0.03  & 0 & 0 & --  & --  & -- & -- & -- & -- & -- \\
ER       & 2000 & 0.03  & 0 & 0 & --  & --  & -- & -- & -- & -- & -- \\ \midrule
DER      &  500 & 0.03 & 0 & 0 & 0.1 & --  & -- & -- & -- & -- & -- \\
DER      & 1000 & 0.03 & 0 & 0 & 0.1 & --  & -- & -- & -- & -- & -- \\
DER      & 2000 & 0.03 & 0 & 0 & 0.1 & --  & -- & -- & -- & -- & -- \\ \midrule
DER++    &  500 & 0.03 & 0 & 0 & 0.3 & 0.5 & -- & -- & -- & -- & -- \\
DER++    & 1000 & 0.03 & 0 & 0 & 0.3 & 0.5 & -- & -- & -- & -- & -- \\
DER++    & 2000 & 0.03 & 0 & 0 & 0.3 & 0.5 & -- & -- & -- & -- & -- \\ \midrule
Refresh  &  500 & 0.03 & 0 & 1e-4 & 0.3 & 0.5 & -- & -- & -- & 1e-5 & 1 \\
Refresh  & 1000 & 0.03 & 0 & 1e-4 & 0.3 & 0.5 & -- & -- & -- & 1e-5 & 1 \\
Refresh  & 2000 & 0.03 & 0 & 1e-4 & 0.3 & 0.5 & -- & -- & -- & 1e-5 & 1 \\ \midrule
STAR     &  500 & 0.1  & 0 & 0 & --  & --  & 0.01 & 0.1 & 1 & -- & -- \\
STAR     & 1000 & 0.1 & 0 & 0 & --  & --  & 0.05 & 0.1 & 1 & -- & -- \\
STAR     & 2000 & 0.1 & 0 & 0 & --  & --  & 0.01 & 0.1 & 1 & -- & -- \\
\bottomrule
\end{tabular}
\end{table*}

\subsection{Generalization to Video Continual Learning}
Tables \ref{tab:acc_vid} and \ref{tab:fr_vid} present the numerical results of average accuracy and forgetting rate, respectively, for iCaRL approach with and without IM on two video datasets (UCF101 and ActivityNet). Combining iCaRL with IM shows improvement in average accuracy and reduces forgetting rate across different memory settings.

\begin{table*}[h]
\centering
\renewcommand{\arraystretch}{1.3}
\vspace{\baselineskip}
\caption{
\textbf{Average Accuracy on UCF101 and ActivityNet.} This table shows
the average accuracy of iCaRL with and without IM across various sizes of memory buffer on two datasets (UCF101 and ActivityNet). The results demonstrate that the integration of information maximization (IM) regularizer consistently outperforms iCaRL on both datasets regardless of the memory setting. 
}\label{tab:acc_vid}
\vspace{\baselineskip}
\begin{tabular}{@{}lcccccc@{}}
\toprule
Dataset & \multicolumn{3}{c}{UCF101} & \multicolumn{3}{c}{ActivityNet} \\ \midrule
Buffer  & 505     & 1010    & 2020   & 1000      & 2000     & 4000     \\ \midrule
iCaRL   & 68.44   & 74.70   & 81.07  & 41.72     & 45.05    & 46.91    \\
iCaRL (IM) & \textbf{75.60} & \textbf{79.11} & \textbf{83.53} & \textbf{45.76} & \textbf{50.01} & \textbf{55.20} \\ \bottomrule
\end{tabular}
\end{table*}
\vspace{\baselineskip}
\begin{table*}[h]
\centering
\renewcommand{\arraystretch}{1.5}
\vspace{\baselineskip}
\caption{
\textbf{Forgetting Rate on UCF101 and ActivityNet.} This table shows
the forgetting rate of iCaRL with and without IM across various sizes of memory buffer on two datasets (UCF101 and ActivityNet). The results demonstrate that the integration of information maximization (IM) regularizer consistently achieves lower forgetting rates on both datasets regardless of the memory setting. 
}\label{tab:fr_vid}
\vspace{\baselineskip}
\begin{tabular}{@{}lcccccc@{}}
\toprule
Dataset & \multicolumn{3}{c}{UCF101} & \multicolumn{3}{c}{ActivityNet} \\ \midrule
Buffer  & 505     & 1010    & 2020   & 1000      & 2000     & 4000     \\ \midrule
iCaRL   & 25.55   & 20.37   & 14.31  & 21.29     & 19.46    & 18.55    \\
iCaRL + IM & \textbf{16.87} & \textbf{16.02} & \textbf{10.54} & \textbf{19.01} & \textbf{18.55} & \textbf{15.48} \\ \bottomrule
\end{tabular}
\end{table*}
\vspace{\baselineskip}
\end{document}